\documentclass{article} 

\usepackage{arxiv}
\usepackage{amsmath,amssymb,amsfonts}
\usepackage{algorithmic}
\usepackage{multirow}
\usepackage{booktabs}

\usepackage{graphicx}
\graphicspath{{./pdf/}{./jpeg/}{./png/}}
\usepackage[T1]{fontenc}
\usepackage[utf8]{inputenc}
\usepackage{nicefrac}
\usepackage{hyperref}
\usepackage{array}
\DeclareGraphicsExtensions{.pdf,.jpeg,.png}

\usepackage{textcomp}
\usepackage{xcolor}
\usepackage{verbatim}
\usepackage{algorithm}
\usepackage{rotating}
\usepackage{float}
\usepackage{stackrel}
\usepackage{subscript}
\usepackage[caption=false]{subfig}
\usepackage{url}
\usepackage{comment}

\usepackage{amsthm}
\usepackage{amsmath}
\theoremstyle{definition}

\AtBeginDocument{%
\abovedisplayskip=-30pt
\abovedisplayshortskip=-10pt
}

\begin{document}

\title{Fair Overlap Number of Balls (Fair-ONB): A Data-Morphology-based Undersampling Method for Bias Reduction}

\author{Jos{\'e} Daniel Pascual-Triana\thanks{Andalusian Institute of Data Science and Computational Intelligence (DASCI), University of Granada, Granada, 18071, Spain.}\> \thanks{Corresponding authors: jdpascualt@ugr.es, alfh@ugr.es.} \\ \and \textbf{Alberto Fern{\'a}ndez}\label{cor2}\footnotemark[1]\> \footnotemark[2] \\ \and \textbf{Paulo Novais} \footnotemark[1]\> \thanks{ALGORITMI CENTRE/LASI, University of Minho, Braga, Portugal.}\>  \\ \and \textbf{Francisco Herrera}\footnotemark[1]\> \thanks{ADIA Lab, Al Maryah Island, Abu Dhabi, United Arab Emirates.}}

\date{Received: date / Accepted: date}
\maketitle

\begin{abstract}

One of the key issues regarding classification problems in Trustworthy Artificial Intelligence is ensuring Fairness in the prediction of different classes when protected (sensitive) features are present. Data quality is critical in these cases, as biases in training data can be reflected in machine learning, impacting human lives and failing to comply with current regulations. One strategy to improve data quality and avoid these problems is preprocessing the dataset. Instance selection via undersampling can foster balanced learning of classes and protected feature values. Performing undersampling in class overlap areas close to the decision boundary should bolster the impact on the classifier. This work proposes Fair Overlap Number of Balls (Fair-ONB), an undersampling method that harnesses the data morphology of the different data groups (obtained from the combination of classes and protected feature values) to perform guided undersampling in overlap areas. It employs attributes of the ball coverage of the groups, such as the radius, number of covered instances and density, to select the most suitable areas for undersampling and reduce bias. Results show that the Fair-ONB method improves model Fairness with low impact on the classifier's predictive performance.


\end{abstract}

\section{Introduction}

The vast amount of data generated in this digital era \cite{hassani_driving_2023}, ranging from simple sensor measurements to biometric data, makes harnessing the relevant information an impossible task unless some processes are automated. Machine Learning techniques 
learn the inner patterns in data, making it possible for computers to predict the behaviour of unseen data \cite{aggarwal_artificial_2021}.  

To learn these patterns and, for instance, the classes in supervised classification problems, the characteristics of each dataset are crucial \cite{luengo_2020}. Models learned from poor-quality data do not reach high predictive performance and might draw unrealistic conclusions \cite{triguero_transforming_2019}. In particular, the presence of class overlap hinders the model's estimation of class boundaries, class imbalance can reduce the importance of some classes in the model (which complicates an equable predictive performance) and noisy samples can distort the predictions. 

When datasets include personal data or sensitive features that might give rise to discrimination, low data quality or mistreatment of said features can cause models to include discriminatory bias \cite{mehrabi_survey_2021}. This is a critical problem nowadays that has led to auditability requirements and the emergence of Artificial Intelligence regulations, such as the recently approved European Union Regulation 2024/1689\footnote{Regulation (EU) 2024/1689 of the European Parliament and of the Council \url{http://data.europa.eu/eli/reg/2024/1689/oj}}, an evolution of the European Union Artificial Intelligence Act\footnote{Artificial Intelligence Act, \url{https://artificialintelligenceact.eu/}}. 

Some sensitive features,  usually referred to as protected features (for example, race or gender), are prone to causing discrimination amongst groups or classes and cannot be included in the model's decision function. However, even if these features are not directly involved in a model, it can still include biases depending on how instances with different values on those protected features are classified (sometimes due to the indirect impact of other highly-correlated features to the protected one that the model does use) \cite{gonzalez_sendino_review_2023}. Data and model Fairness are achieved when these possible biases amongst groups are minimised or eliminated.

While Artificial Intelligence often focuses on optimising prediction model parameters, another approach is focusing on data quality, as models learned on high-quality data tend to achieve high performance \cite{luengo_2020}. To reduce model bias, a widespread approach is to modify the dataset via sampling \cite{salazar_fawos_2021}. Sampling data of specific groups or classes, either by adding instances of those with fewer data (oversampling) or selecting for usage only some from the over-represented ones (undersampling), can foster an equitable treatment of all groups or classes by the model. When protected features are present, this would mean balancing the model's performance for instances of various values of the protected features. Specifically, this means balancing the groups formed by the combinations of the class values and protected feature values as, when taken into account separately, not only might bias reduction be harder, but counterproductive situations might arise (for example, if more $gender=female$ samples were added to balance $gender$ but they were added in negative class areas, discrimination against $gender=female$ would increase).

Sampling strategies in the literature follow varied strategies. Most are based on sampling until group balance is reached and, while some strategies do it randomly \cite{rancic_investigating_2021}, others include restrictions or preferences, such as only balancing groups close to the model's decision boundaries \cite{salazar_fawos_2021} or using particular clusters \cite{he_novel_2023}. However, just balancing a priori ratios of protected feature values and classes does not always yield equal performance of the model (particularly, when the distribution of instances with each protected feature value is different).

The hypothesis of this work is that, to balance model performance on different data groups, the focus must be on the group overlap areas of the problem, as those would be the sections of the data space were sampling would be the most effective at modifying model decisions and, thus, at improving Fairness. To this aim, using data morphology, which has proven to be a remarkable strategy for the characterisation of groups or classes and the detection of overlap amongst them, would be a well-founded approach.

Amongst the strategies that locate and measure overlap in tabular datasets, the use of "Overlap Number of Balls" (ONB) is proposed \cite{pascual-triana_revisiting_2021}. This technique is based on the coverage of data using balls that only include points from a single group. ONB employs morphology to characterise the different groups in data, as well as to estimate group overlap and classifier performance. Furthermore, the information on the boundaries amongst classes or groups provided by the ball coverage can also be harnessed for model explainability \cite{pascual_triana_overlap_2024}.

The aim of this work is to leverage the overlap estimation capabilities of ONB and the characterisation of groups (obtained using class and protected feature value combinations) derived from its coverage strategy  to improve Fairness in classification models according to protected feature values without a noticeable reduction in model performance. To this end, we propose a preprocessing method via sampling, called Fair Overlap Number of Balls (Fair-ONB), that is based on group morphology and data neighbourhoods. Since undersampling allows for the reduction of both bias and noise (particularly due to using morphology to locate it) and oversampling would only tackle the former, the proposed strategy will perform undersampling. 

The Fair-ONB method involves two stages: selecting the undersampling groups, and then using percentile thresholds on coverage ball attributes, such as ball radius, number of instances covered and density, to decide how to undersample. It produces stable results, compared to random sampling strategies \cite{rancic_investigating_2021, valentim_impact_2019}, due to using empirical measures to decide which samples to eliminate, and the location of balls under the chosen thresholds (and the change in group size that this would entail) can be easily quantified. 

The efectiveness of the proposed Fair-ONB method will be tested on several of the most broadly-used real datasets in data Fairness studies. The a priori behaviour of the different threshold variables for instance filtering is studied, both in terms of predictive performance and Fairness
. The results are compared with FAWOS, a state-of-the-art, neighbourhood-based sampling technique \cite{salazar_fawos_2021}.

The rest of this paper is organised as follows. Firstly, Section \ref{prelim} presents the preliminaries on Fairness, discrimination and bias. Then, Section \ref{subsec:Fair-ONB-met} details the proposed Fair-ONB method. Next, Section \ref{frame} describes the experimental framework for the evaluation of the Fair-ONB method's behaviour. Section \ref{exper} shows the experimental results. Lastly, Section \ref{concl} presents the concluding remarks from this manuscript and some future work lines.

\section{Preliminaries on Discrimination and Bias}\label{prelim}

This Section presents some preliminaries regarding bias in Machine Learning. Section \ref{prelim_fair_metr} introduces important concepts and some of the different notions of Fairness, along with their associated metrics. Section \ref{prelim_bias_red} shows some of the techniques present in the state of the art for the minimisation of bias, with an emphasis on sampling techniques.

\subsection{On Fairness metrics}\label{prelim_fair_metr}

The existence of discrimination amongst groups of people, that is, the treatment that inherently favours some of them, involves substantial ethical problems and is legally unacceptable\footnote{Regulation (EU) 2024/1689 of the European Parliament and of the Council \url{http://data.europa.eu/eli/reg/2024/1689/oj}}. Fairness in Machine Learning means the absence of bias or discrimination amongst groups of data. This Fairness is checked with respect to protected features, such as gender or race, which are variables in datasets whose unequal treatment would expose the existence of discrimination.

Fairness and bias can be measured in different ways according to the situation, as studied in multiple surveys  \cite{mehrabi_survey_2021, pagano_bias_2023,  caton_fairness_2024}.
For example, it might be preferable that an equal rate of acceptance for a grant between men and women but, when assigning a medical procedure, the actual need for such treatment (which might be different according to gender) must also be taken into account.

According to these preferences, there are multiple possible approaches in order to determine if a classifier is biased \cite{pagano_bias_2023}. For example, in some cases the aim is to maintain demographic parity, also named statistical parity, which involves the positive prediction being independent on the protected feature; that is, given a prediction Ŷ, demographic parity is measured according to how different the chances of obtaining a prediction Ŷ=1 is for each value of protected feature P. This would create 2 similar metrics:

\begin{itemize}
    \item \textbf{Statistical Parity Difference (SPD)}: it measures the difference between the probabilities of a positive classification between two protected feature values (Equation \ref{dpe}). In the optimum case (equal chances), its value is 0.

\begin{equation}
SPD =Pr(\widehat{Y}=1 |P=0)-Pr(\widehat{Y}=1 |P=1)\label{dpe}
\end{equation}

    \item \textbf{Disparate Impact (DI)}: it measures the ratio between the probabilities of a positive classification between two protected feature values (Equation \ref{id}). In the optimum case (equal chances), its value is 1. Given its extended use to check biases in the USA, this metric is commonly applied \cite{caton_fairness_2024}.

\begin{equation}
DI=\frac{Pr(\widehat{Y}=1 |P=0)}{Pr(\widehat{Y}=1 |P=1)}\label{id}
\end{equation}
    
\end{itemize}

In some cases, a different version of Disparate Impact is used, named Adapted Disparate Impact (ADI), whose values are between 0 and 1. Its formula is given in Equation \ref{ida}.

\begin{equation}
ADI=\left\{ \begin{array}{lr} ID & : ID \leq 1\\ \frac{1}{ID} & : ID > 1 \end{array} \right.\label{ida}
\end{equation}

Other extended metrics would involve equality of probability (also named equalised odds) and equality of opportunity, whose foci are the true positive and/or false positive rates being the same for different values of the protected feature. Their associated metrics are measured according to the difference in probabilities for each group, and are named Equal Probability Difference (EPD) and Equal Opportunity Difference (EOD), whose formulae are given in Equations \ref{dip} and \ref{dio}, respectively.

\begin{equation}
EPD=Pr(\widehat{Y}=1 |Y,P=0)-Pr(\widehat{Y}=1 |Y,P=1)\label{dip}
\end{equation}

\begin{equation}
EOD=Pr(\widehat{Y}=1 |Y=1,P=0)-Pr(\widehat{Y}=1 |Y=1,P=1)\label{dio}
\end{equation}

Unless they are based on the same types of probabilities (such as in SPD and DI), optimizing multiple bias metrics simultaneously is generally unfeasible \cite{kleinberg_inherent_2017}; therefore, studies often select one of them to measure Fairness.

\subsection{Bias reduction strategies}\label{prelim_bias_red}

Having indicated multiple metrics that measure Fairness, knowing how to reduce biases is also important. The use of model bias reduction techniques so that they comply with the current Fairness regulations is a currently thriving research area \cite{gonzalez_sendino_review_2023}. Consequently, there are many techniques for bias reduction, from adding Fairness restrictions to the model's objective function \cite{wu_multi_objective_2022}, to adversarial training \cite{reimers_conditional_2021}, adding weights to data with specific protected feature values \cite{yu_fairbalance_2024} or modifying the dataset so it includes less bias before learning the model \cite{salazar_fawos_2021}. Since this study is focused on dataset modifications via sampling, the strategies in that area will now be outlined.

Most sampling strategies are mainly based on the equal treatment of the subgroups of data defined by the combinations of class and protected feature values (for example, the pair of $Class=1$ and $Gender=F$). This is due to Fairness metrics taking both into account simultaneously and, thus, seeking only class balance or protected feature balance on their own would not be sufficient. 

According to when these classifier bias reducing techniques are applied, they can be divided into 3 groups: preprocessing strategies, which work on the base data; inprocessing strategies, that work during the learning process of the model; or postprocessing strategies, that correct the model's results after its prediction.

\textbf{Preprocessing} strategies work on the training data that will be used to learn the classifier. The aim of these techniques is to detect whether the dataset includes a priori anomalies or biases amongst the different subgroups in order to modify it appropriately \cite{kamiran_data_2012}. This way, the models would not learn and reflect the biased behaviours stemming from data collection.

This a priori bias reduction so that the posterior model can have similar learning capabilities for every group is usually obtained by balancing the number of samples for each subgroup (according to the combination of classes and protected feature values) in the dataset. To this end, the number of samples of the over-represented groups can be reduced (undersampling), or it can be increased for the under-represented groups (oversampling). This sampling is often done randomly in the appropriate subgroups \cite{rancic_investigating_2021, valentim_impact_2019}, but sometimes specific preferences are introduced (such as randomly sampling on the model's boundaries \cite{salazar_fawos_2021, sun_trade_off_2022}, or amongst clusters \cite{chakraborty_bayesian_2020, he_novel_2023, sonoda_fair_2023}) or samples are selected iteratively \cite{anahideh_fair_2022}. Another preprocessing technique consists on the modification of data labels (massaging) of certain bias-inducing samples \cite{sharma_data_2020}. Albeit another technique is the elimination of the protected feature so that the model is not directly exposed to bias, this strategy does not always work as intended \cite{valentim_impact_2019}, since other features with high correlations with that protected feature can still be present and, thus, the influence of that same protected feature would be indirectly maintained; this would be the case, for example, of post code and race or wage.

Amongst preprocessing techniques, it would be important to highlight Fairness-AWare OverSampling (FAWOS) \cite{salazar_fawos_2021}, a sampling strategy based on instance neighbourhoods. FAWOS employs the k nearest neighbours method to label instances as safe (4-5 of its neighbours share its class), borderline (2-3 of its neighbours share its class), rare (only 1 neighbour shares its class) and outlier (no neighbours share its class). The labels of subgroups with positive class and protected feature values that are discriminated against are used to grant different probabilities for those instances for Synthetic Minority Over-sampling Technique (SMOTE) usage, thus balancing the groups.

Regarding \textbf{inprocessing} and \textbf{postprocessing} techniques, the learning or modification of the model can allow it to work with biased datasets directly. In these cases, different weights can be assigned to data from different groups, so that model learners would favour those that would otherwise be disadvantaged \cite{krasanakis_adaptive_2018, zeng_fairness_aware_2023}. Constraints can also be applied so that the model is learned taking Fairness into account \cite{ferry_improving_2023, liu_sampling_2021}. Another strategy would be using multiobjective optimisation, including both Fairness and performance at the same time \cite{anahideh_fair_2022}, or using adversary learning where certain protected feature values are swapped \cite{zeng_fairness_aware_2023, zhang_automatic_2022}. Other techniques include a posteriori massaging of labels \cite{he_novel_2023} or interaction with experts \cite{anahideh_fair_2022, wang_interactive_2021}.

\section{The Fair Overlap Number of Balls method}\label{subsec:Fair-ONB-met}

This Section describes in detail the functioning of the Fair-ONB method, the novel undersampling proposal based on data morphology. This method undersamples groups with favourable bias so that protected groups are treated similarly by the classifier. This is done in a guided manner, harnessing the morphology of the group coverage obtained from the Overlap Number of Balls (ONB) algorithm and its ball properties in order to select problematic areas of the data space.

Firstly, the ONB algorithm is explained in Section \ref{ONB}. Then, some preliminaries on the choices made during the method's creation are indicated in Section \ref{prelim-fair-onb}. Lastly, the structure and inner working of the method are introduced in Section \ref{estruc-fair-onb}.

\subsection{Overlap Number of Balls}\label{ONB}

ONB is a class boundary estimation method that was initially conceived as a complexity metric, that is, a technique that measures how difficult to classify a dataset is according to its inner characteristics \cite{pascual-triana_revisiting_2021}. ONB is based on the Pure Class Cover Catch Digraph (P-CCCD) coverage algorithm \cite{manukyan_classification_2016}. 

\begin{figure}[ht]
    \centering

    \subfloat[]{\label{ejbolasa}
    \includegraphics[width=.45\linewidth]{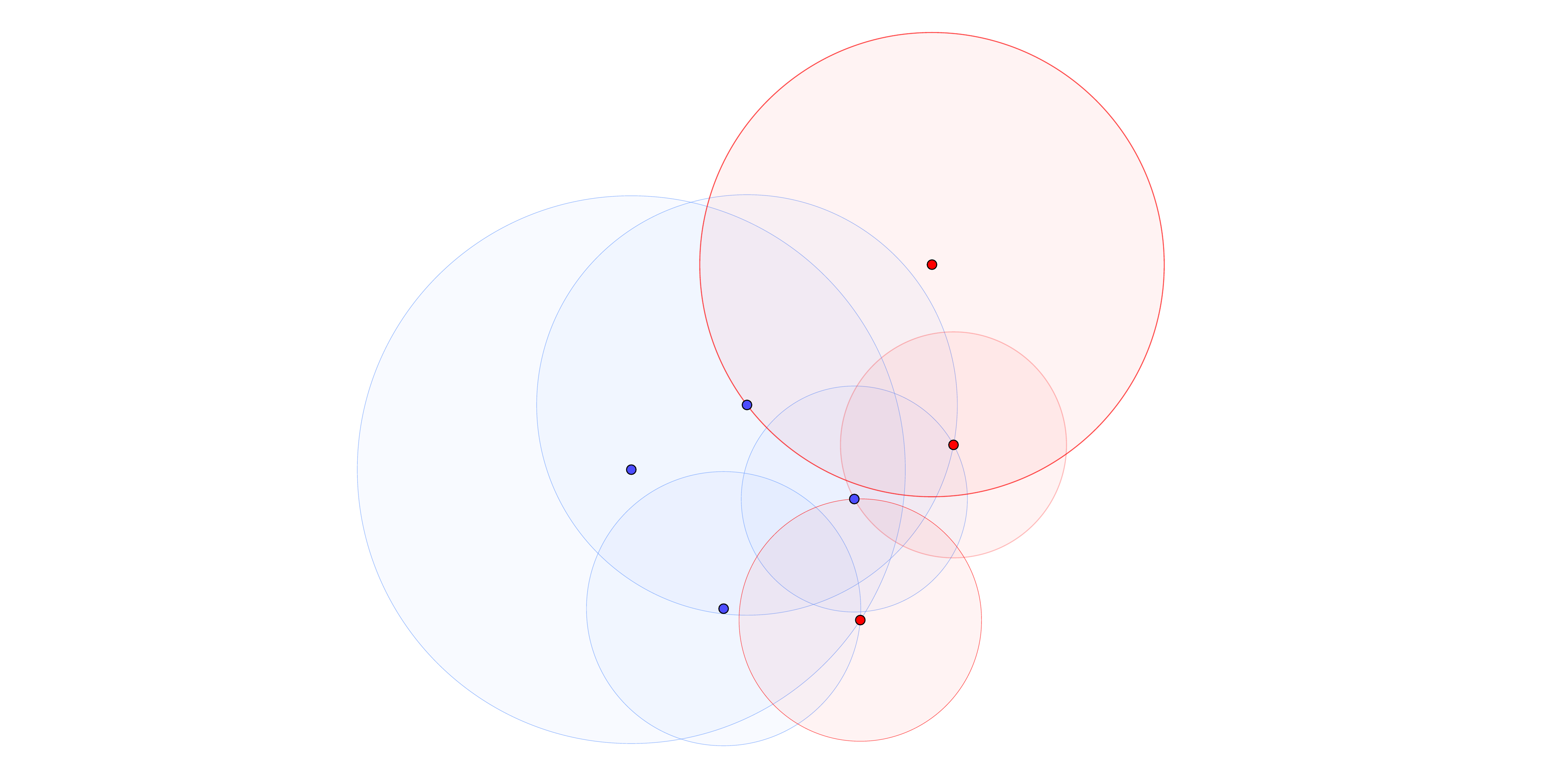}}
    \subfloat[]{\label{ejbolasb}
    \includegraphics[width=.45\linewidth]{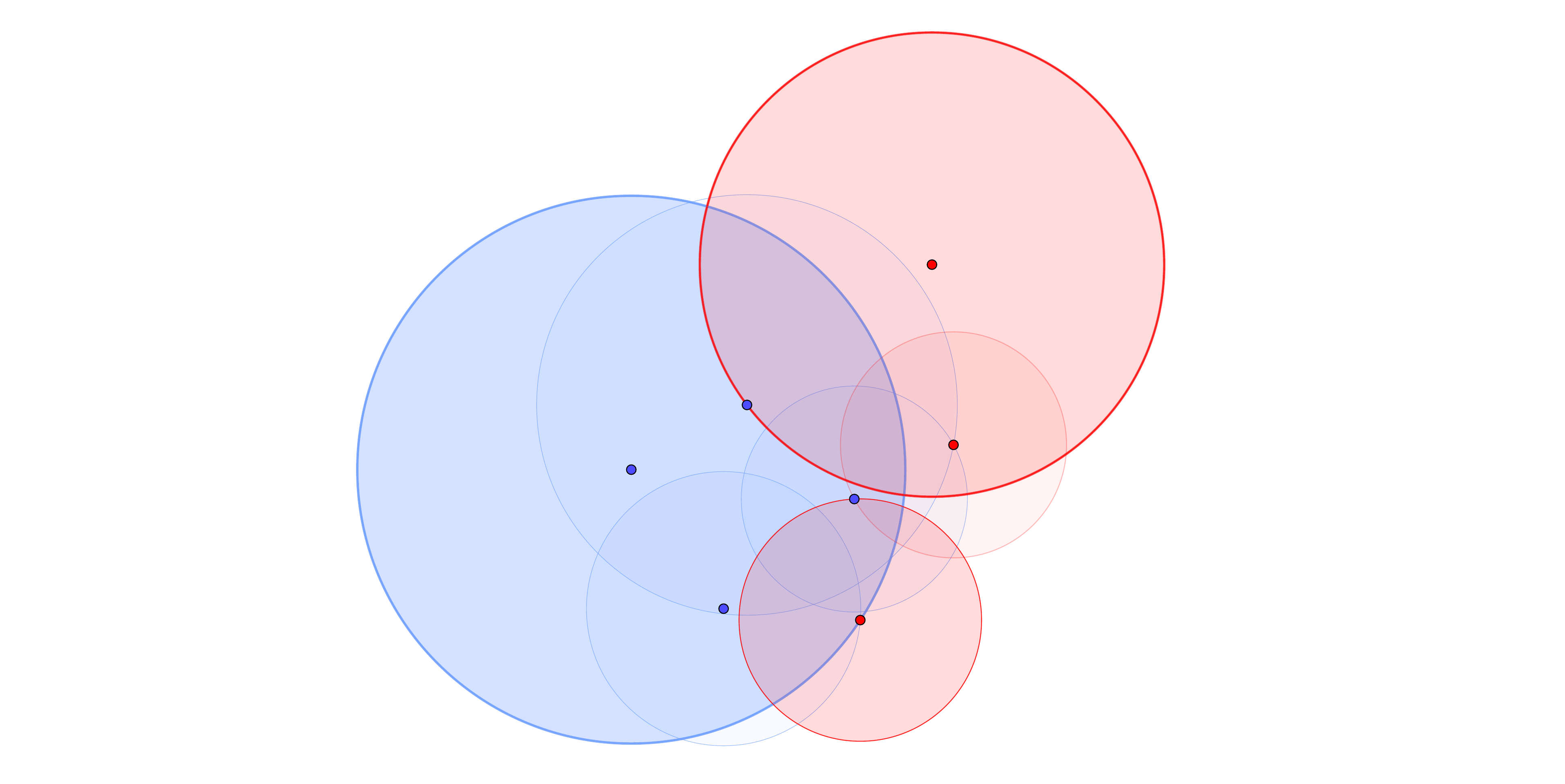}}

\caption{\label{diagrama_ONB}Example of the P-CCCD ball coverage algorithm that serves as the base for ONB. Firstly, the maximum radii for the balls centred on each instance that do not include points of other classes are calculated (\ref{ejbolasa}), and, afterwards, the balls that cover the most points of their class are iteratively selected until all points are covered  (\ref{ejbolasb}).}
\end{figure}

Its strategy is simple. Firstly, for each instance in the dataset, the biggest open ball (the interior of the hypersphere of the same radius) that does not cover any points of a different class is generated. Then, for each class, the balls that would cover the most points that are not yet covered by other balls in the coverage set are iteratively chosen as part of said coverage set, until all points of that class are covered. An example of this is shown in Figure \ref{diagrama_ONB}.

This coverage strategy has very useful implications. 
\begin{itemize}
\item The coverage locates and maps classes in the dataset. This estimation of the class distribution gives information on where it would be normal to find points of each class, and where their presence could be considered unusual or noisy.
\item The number of balls that are necessary to cover the dataset gives information on both how difficult to classify the dataset is and the degree of overlap amongst classes. This information is very useful towards deciding whether preprocessing is necessary in order to improve classifier performance \cite{pascual-triana_revisiting_2021}.
\item The ball coverage can estimate class boundaries, either in the base dataset or in the classifier that is learnt from it. In the latter case, the morphology of these estimated boundaries can be used to improve the explainability of complex classification models. In particular, Overlap Number of Balls Model-Agnostic Counterfactuals (ONB-MACF) is a counterfactual generator that employs the boundary estimations of ONB to provide counterfactual explanations \cite{pascual_triana_overlap_2024}.
\end{itemize}

\subsection{Preliminaries on the Fair-ONB method}\label{prelim-fair-onb}

Given a dataset whose protected features and class are known, the combination of their values is employed to detect and correct biases, given their paired involvement in Fairness estimation formulae. For the Fair-ONB method, each of these combinations forms a group. For example, a dataset with binary protected features ``race'' and ``gender'' and a binary class would have the groups shown in Table \ref{Fair-ONB-grupos}).

\begin{table*}[htbp]
\caption{Groups obtained from the combinations of protected feature and class values.}
\begin{center}
\begin{tabular}{cccccccccc}
\hline
\textbf{Race}&\textbf{Gender}&\textbf{Class}&\textbf{Group} \\
\hline
0 & 0 & 0 & 0\\

0 & 0 & 1 & 1\\
0 & 1 & 0 & 2\\
0 & 1 & 1 & 3\\
1 & 0 & 0 & 4\\
1 & 0 & 1 & 5\\
1 & 1 & 0 & 6\\
1 & 1 & 1 & 7\\

\hline

\end{tabular}
\label{Fair-ONB-grupos}
\end{center}
\end{table*}

The aim is to see how those groups are distributed, since, depending on their overlap and their degrees of imbalance, the classifier learned using the dataset can be biased.

At the same time, before employing the method, the existence of a priori biases in the dataset or model is checked using a Fairness metric. In this study, said Fairness metric is the Disparate Impact (DI, Equation \ref{id}), which, as a reminder, is the ratio between the probabilities of obtaining a positive result  (Ŷ=1) according to the values in the protected feature. This first check allows the selection of which subgroups need to be preprocessed.

\begin{figure}[!h]
\centerline{\includegraphics[width=0.45\linewidth]{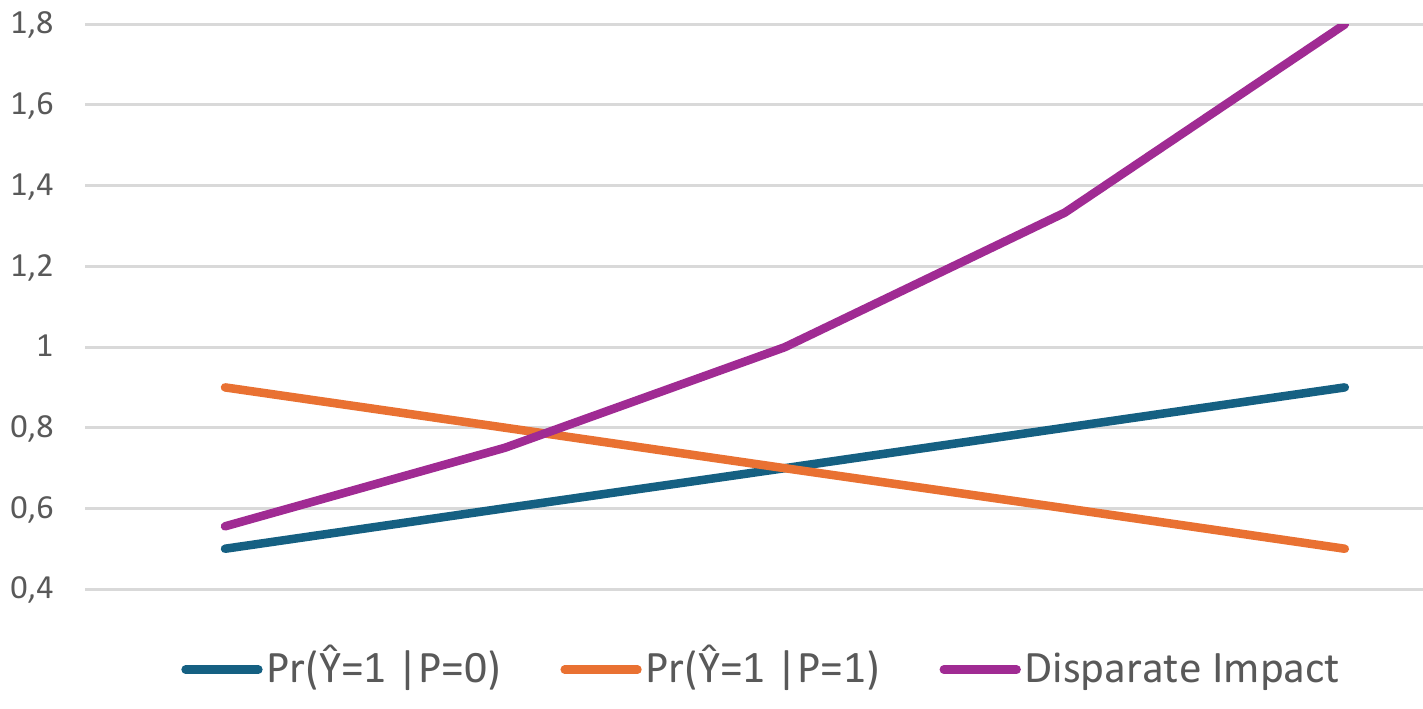}}
\caption{Disparate impact plot as the preferences of the classifier change.}
\label{fig_ejemplo_id}
\end{figure}

As shown in Equation \ref{id}, when no bias is present according to a protected feature, $DI=1$; otherwise, the value is still considered good when it is between 0.8 and 1.25 \cite{caton_fairness_2024}. Figure \ref{fig_ejemplo_id} shows the behaviour of DI as the preferences of the model change (as happens when its training data is modified). It includes the neutral DI value (1) and the positive classification probabilities for the different values of a protected feature.

With this in mind, DI can be checked on the base dataset (with the existing ratios, to detect pre-existing bias in the dataset) or on the model (using the predictions given by the classifier, which elucidates whether the model is biased) for each of the protected features. When $DI>1$, the bias favours the group with protected feature value 0; when $DI<1$, it is biased towards the feature value 1.

\subsection{Structure and inner working of the Fair-ONB method}\label{estruc-fair-onb}

The Fair-ONB method is based on the characterisation of groups derived from the combination of class and protected feature values in the dataset. This is done via the data coverage using balls obtained from ONB (see Section \ref{ONB}). 

Once the coverage is obtained, the Fair-ONB method undersamples instances of favoured groups to foster a fair classification. This is done in a guided way, selecting instances in problematic areas of the data space after harnessing the properties of coverage balls.

The Fair-ONB method selects the balls for undersampling in two stages: first, the groups whose balls can be selected for undersampling are chosen; then, using their characteristics, some balls of those groups (and the instances they contain) are eliminated.

\paragraph{Stage 1: group selection for preprocessing.}

When Fairness metrics indicate bias towards a value of a protected feature in the dataset (which would likely translate into that same bias being learnt by the subsequent classifier), the chosen strategy is to perform undersampling of the groups with positive class (Ŷ=1) and said value (either 0 or 1) in the protected feature. Since there can be multiple protected features simultaneously, there would be two ways of choosing the groups to preprocess.

\begin{itemize}
    \item Using the union of the groups with positive class and favoured values in the different protected features (OR). Thus, if in the example of Table \ref{Fair-ONB-grupos} there are biases favouring $race=1$ and $gender=0$, the groups to be preprocessed would be groups 1, 5 and 7.
    \item Using the intersection of the groups with positive class and favoured values in the different protected features (AND). Likewise, if in the example of Table \ref{Fair-ONB-grupos} there are biases favouring $race=1$ and $gender=0$, only group 5 would be chosen for preprocessing.
\end{itemize}

Depending on the dataset characteristics, the chosen undersampling strategy can greatly affect the results. For that reason, both group selection methods will be studied.

\paragraph{Stage 2: selection of balls from those groups for elimination.}

Once the ball coverage has been obtained, some ball attributes that might indicate noisy areas or group overlap that can favour the existence of bias are studied. In particular, the radius, the number of instances they cover and the density of balls will be observed. The hypotheses regarding their possible effects on data bias are now described.

\begin{itemize}
    \item \textbf{Radius}: if it is very small, the ball is in an area of group overlap, be it due to data distribution or to the existence of noise. 
    \item \textbf{Number of covered instances}: if it covers very few instances, the ball might not be truly representative in the characterisation of its class. It can also be in an area of group overlap or noise.
    \item \textbf{Density}: if it is very low, the ball might have been generated using an isolated point far from the decision boundary or just very far inside a class' area, which might not contribute much to the model's decisions. While some balls with very high density can also be problematic (due to, for example, having a very small radius), these particular cases are already covered by the other premises, so highly dense but useful balls will not be affected.
\end{itemize}

To perform the undersampling, percentiles (values of an attribute for which exactly that percentage of cases have a lower value) of those three characteristics in the coverage are calculated, and will serve as thresholds for ball selection. Using percentiles instead of predefined values aims to have a better adaptability to each dataset.

Once obtained, the thresholds of all three ball characteristics are combined to decide different degrees of preprocessing (for example, one test can simultaneously use percentile 5 for radius, 15 for number of covered instances and 10 for density). The balls from the selected preprocessing groups whose radii, number of covered instances or densities are lower than each test's chosen ball characteristic thresholds are eliminated, along with the instances they cover, due to being considered problematic.

\begin{figure}[!h]
\centerline{\includegraphics[width=0.9\linewidth]{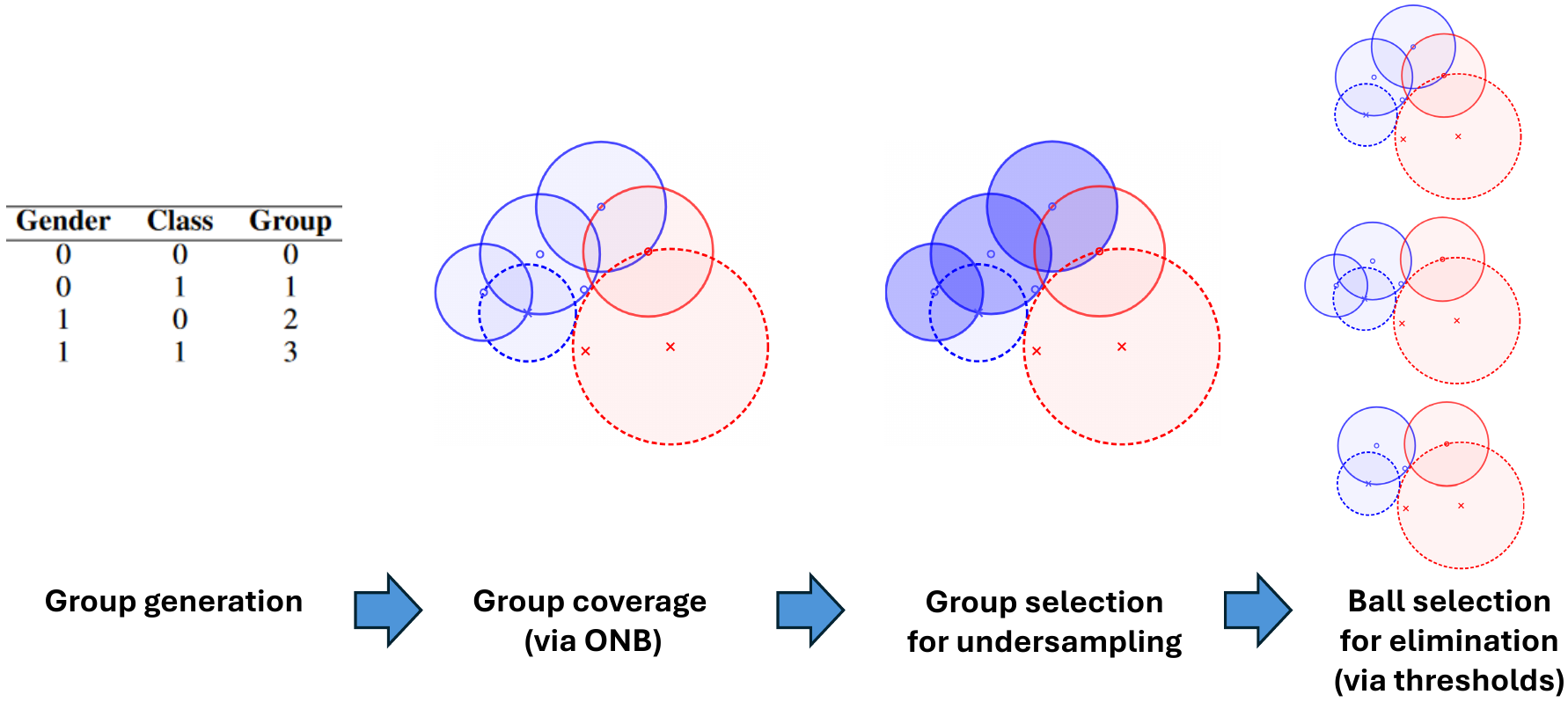}}
\caption{The overall strategy of the Fair-ONB method starts with the generation of the groups defined by the pairs of class and protected feature values. The group coverage is then performed using ONB (in the plot, color is used to discern protected feature values, and instance symbols and circumference design to indicate class values). According to a chosen Fairness metric, groups are selected for undersampling. Finally, undersampling is performed, eliminating balls of that group employing thresholds (in the example, radius, density and number of covered instances were employed, respectively, from top to bottom).}
\label{fig_overall}
\end{figure}

The overall modus operandi of the Fair-ONB method is shown in Figure \ref{fig_overall}.

\section{Experimental Framework}\label{frame}

This Section presents the experimental framework that tests the goodness of the proposed Fair-ONB method, with the aim of improving model Fairness while maintaining performance. Section \ref{subsec:Fair-ONB-exp-conj} presents the datasets that will be used in the experiments. Section \ref{subsec:Fair-ONB-exp-estruc} explains the structure of the experiments and their parameters.

\subsection{Datasets}\label{subsec:Fair-ONB-exp-conj}

This experimental framework includes the use of four datasets. The first three (COMPAS, Adult and German) are the most used in the field of bias reduction \cite{fabris_algorithmic_2022}; the fourth dataset, Ricci, was included to add more broadness to the study.

The exact versions of the aforementioned datasets have been modified to reduce the computational complexity: categorical variables were transformed into binary ones via one-hot encoding and, in the case of Adult, its size was cut in half (while maintaining its structure, via stratified sampling). The dataset characteristics (after those transformations) are as follows.

\begin{itemize}

\item The COMPAS dataset predicts whether convicts will reoffend in the subsequent two years, according to their personal data and criminal history. Once transformed, it includes 6172 samples, 7 attributes (3 of them binary, 4 numeric) plus the binary class. The protected features are race and sex.

\item The Adult dataset evaluates whether a person will earn over 50.000\$/year according to census data. Once transformed, the reduced dataset includes 24.416 samples, 13 attributes (7 of them binary, 6 numeric) plus the binary class. The protected features are race and sex.

\item The German dataset predicts the credit risk of people. Once transformed, it includes 1.000 samples, 24 features (14 of them binary, 10 numeric) plus the binary class. The protected features are gender and age.

\item The dataset Ricci evaluates the promotion of a group of firefighters according to exam results. It includes 118 samples, 4 attributes (2 of them binary and 2 numeric) plus the binary class. The protected feature is race.

\end{itemize}

\subsection{Experimental structure and parameters}\label{subsec:Fair-ONB-exp-estruc}

This study has two parts, each of them with a different aim.

\begin{enumerate}
    \item Testing the behaviour of the different threshold variables (according to their percentiles) in different situations, to detect whether there are certain thresholds that generally give good results.
    \item Comparing the bias reduction when using the Fair-ONB method to that of FAWOS \cite{salazar_fawos_2021}, a recent neighbourhood-based sampling strategy with whose strategy Fair-ONB shares some similarities.
\end{enumerate}

In order to properly evaluate classifier performance, 5-fold cross validation is used in each experiment for each dataset. Sampling methods are only applied on each experiment's training sets. The predictive performance is evaluated over the test set using two usual metrics: AUC and accuracy. AUC is used in the first part of the study, due to being a more conservative metric and its aptitude in case of imbalance; accuracy is used in the second part, once the good behaviour of the Fair-ONB method has been proven, and to allow for fair comparison with FAWOS (which samples groups until they are approximately even, a situation where accuracy is a useful metric). Fairness is evaluated using Disparate Impact. Performance and Fairness results are aggregated via the arithmetic mean of the 5 folds'. 

For dataset preprocessing using the Fair-ONB method, up to the 5 first different percentiles are used for each threshold feature (radius, number of covered instances, density), starting with 0 and going up in increments of 5. The combination of percentiles for each threshold feature parameterise the different experiments for each dataset. Moreover, there experiments are performed both using union and intersection for undersampling group selection.

The classifier used is a decision tree with its default parameters in the version 0.23.2 of scikit-learn\footnote{\url{https://scikit-learn.org/stable/modules/generated/sklearn.tree.DecisionTreeClassifier.html}} (the necessary version for FAWOS), except for including a fixed seed (\textit{random\_state=30}). These classifiers are learned using the preprocessed training set of each experiment.

In the second part of the study, the Fair-ONB method is compared to FAWOS \cite{salazar_fawos_2021}. The fact that FAWOS is also a sampling method based on sample neighbourhoods makes it the closest strategy to Fair-ONB. FAWOS employs 5NN to label samples as safe, borderline, rare and outlier according to how many of their neighbours share their class, and then uses those labels to allocate probabilities of those samples being selected for SMOTE.

The parameters used for FAWOS are the same as in the original study \cite{salazar_fawos_2021}. The weights for each label group were 0 for safe, 0.6 for borderline and 0.4 for rare points (outliers always get null weight), and oversampling factors were either 0.8, 1 or 1.2.

\section{Experimental results}\label{exper}

This Section presents the experimental results of the two studies. Firstly, Section \ref{subsubsec:Fair-ONB-ana-umbral} evaluates the effects of selecting different undersampling thresholds; then, Section \ref{subsubsec:Fair-ONB-ana-compar} presents the comparison of the results using Fair-ONB and FAWOS, two bias reduction sampling methods.

\subsection{Study of the behaviour of threshold features in the Fair-ONB method}\label{subsubsec:Fair-ONB-ana-umbral}

This Section presents the results of the experiments on threshold combinations. Disparate Impact is measured according to the different protected variables of each of the four datasets, as well as classifier performance according to AUC, for each preprocessing parametrisation. Thus, the aims are to check whether Fairness can be improved without much effect on AUC, and whether certain thresholding parameters can be widely considered useful.
Two sets of experiments are performed on each dataset, depending on whether union or intersection is used for undersampling group selection. 

To better exemplify the different observed behaviours of union and intersection undersampling, the results on COMPAS and Adult will be plotted.  To facilitate a global comparison, in this type of figure, Disparate Impact values for both protected variables in the dataset and AUC values are shown simultaneously; furthermore, to better reflect whether the method improves the results, the base results (with no preprocessing) are shown as a horizontal black line and the optimum Disparate Impact (1) as a green one. In each case, radius will be used as the X axis feature, due to providing the most illustrative results.

\begin{figure}[!h]
\centerline{\includegraphics[width=.9\linewidth]{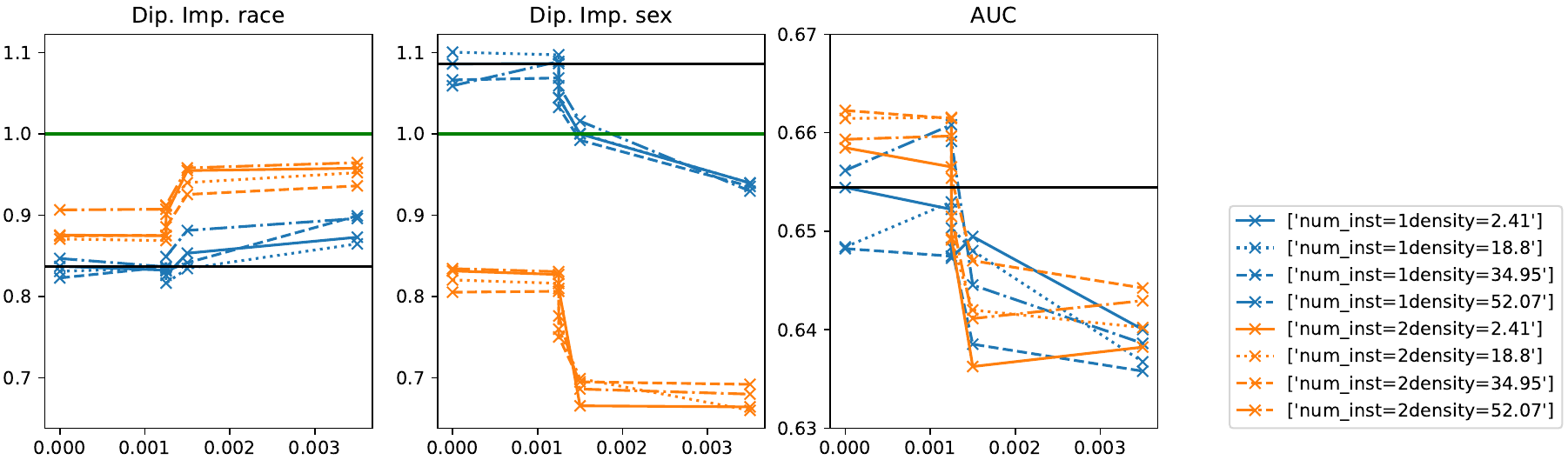}}
\caption{Behaviour of Disparate Impact according to the protected features (race and sex) and of AUC when the radius threshold is modified and the other threshold variables are fixed, using union undersampling on COMPAS.}
\label{compas-union}
\end{figure}

\begin{figure}[!h]
\centerline{\includegraphics[width=.9\linewidth]{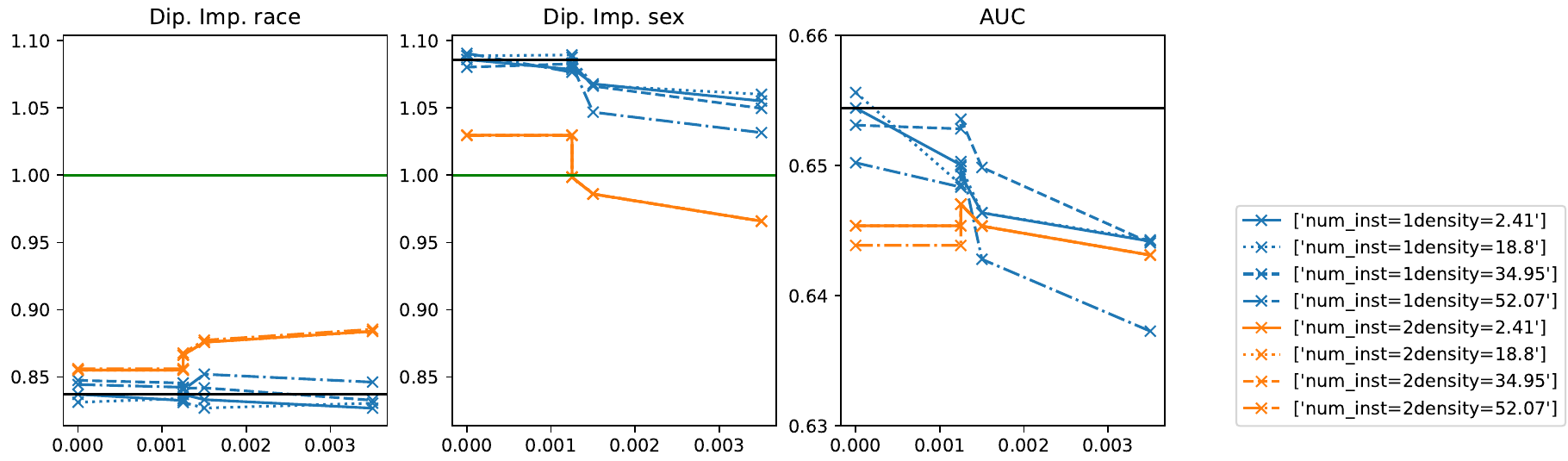}}
\caption{Behaviour of Disparate Impact according to the protected features (race and sex) and of AUC when the radius threshold is modified and the other threshold variables are fixed, using intersection undersampling on COMPAS.}
\label{compas-intersec}
\end{figure}

\paragraph{COMPAS dataset.} Figures \ref{compas-union} and \ref{compas-intersec} respectively depict the union and intersection undersampling results on this dataset. As can be observed, the results when using different thresholds for the number of covered instances create result clusters regarding Disparate Impact, with noticeable differences. Meanwhile, an interesting behaviour is observed for intermediate values of radius thresholds, where Disparate Impact varies greatly in this dataset. Substantial Disparate Impact improvements can be attained on COMPAS (to the point of reaching the optimum on ``sex'' with both types of undersampling, although with different parameters). AUC, while slightly reduced, is comparable to the base value.

\begin{figure}[!h]
\centerline{\includegraphics[width=.9\linewidth]{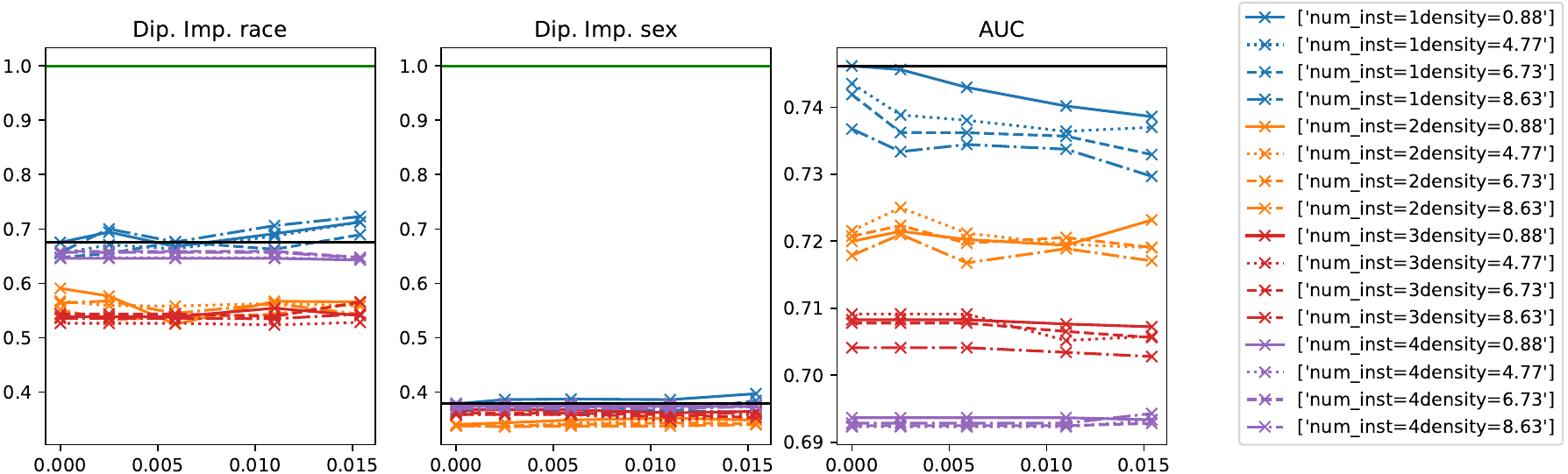}}
\caption{Behaviour of Disparate Impact according to the different protected features 
(race and sex) and of AUC when the radius threshold is modified and the other threshold 
variables are fixed, using union undersampling on Adult.}
\label{adult-union}
\end{figure}

\begin{figure}[!h]
\centerline{\includegraphics[width=.9\linewidth]{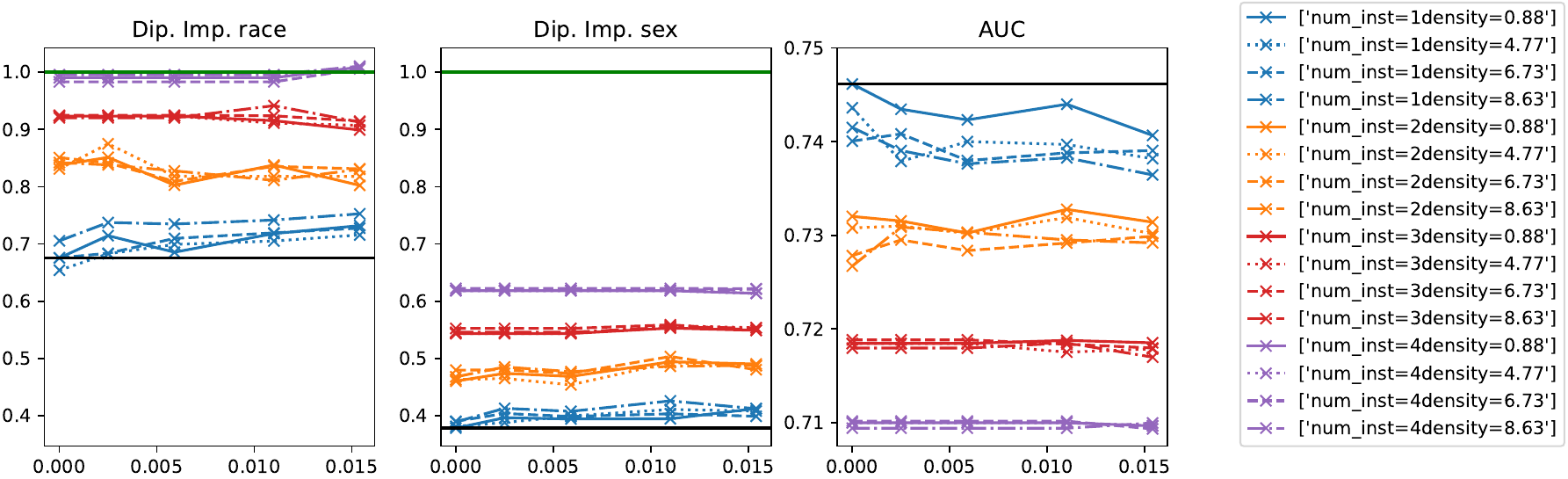}}
\caption{Behaviour of Disparate Impact according to the different protected features 
(race and sex) and of AUC when the radius threshold is modified and the other threshold 
variables are fixed, using intersection undersampling on Adult.}
\label{adult-intersec}
\end{figure}

\paragraph{Adult dataset.} Figures \ref{adult-union} and \ref{adult-intersec} show the union and intersection undersampling results on this dataset. Here, union undersampling does not obtain noticeable improvements. Only slightly better Disparate Impact results were obtained for both protected variables. However, with intersection undersampling, Disparate Impact provides considerable improvements for both protected variables, reaching the optimum in ``race''. In the case of ``sex'', much better results than the baseline are also obtained, while not reaching the often recommended [0.8,1.25] Disparate Impact interval.

\paragraph{Other datasets.}
For the \textbf{German dataset}, only slight improvements were obtained on ``gender'', since it was already very close to the optimum Disparate Impact, while noticeable improvements were attained according to protected feature ``age''. 
In this dataset, like in Adult, very similar global Disparate Impact results were obtained when using union and intersection undersamplings, and in both cases it was possible to improve both Fairness and performance simultaneously.
As for the \textbf{Ricci dataset}, there is only one protected feature (``race''), so it makes no sense to separate union and intersection (since there is only one possible group to undersample). Slight improvements were obtained compared to the base case.

\paragraph{Lessons learned.} As a summary of the results obtained with all datasets, including the best results of each type of undersampling (and the base case for comparison) are presented in Table \ref{compar_german_base2}. For each dataset, the best results on each individual metric are presented in bold, while the results of the parametrisation that minimises DI globally (that it, the one that minimises the total distance to the optimum disparate impacts amongst the protected features) are presented in bold and italics.

\begin{table*}[!h]
\caption{Comparison of Disparate Impact and AUC between the base COMPAS, Adult, German and Ricci datasets and the best
union and intersection undersamplings, along with the best undersamplings for the optimisation of each individual metric.}
\begin{center}
\scalebox{.8}{
\begin{tabular}{ccccccccc}
\hline
\textbf{Dataset}&\textbf{US Type}&\textbf{Aim}&\multicolumn{3}{c}{{\textbf{US Thresholds}}}&\multicolumn{2}{c}{{\textbf{DI by protected feature}}}&\textbf{Perf.} \\
\hline

&&&\textbf{N\_i}&\textbf{R}&\textbf{D}&\textbf{Race DI}&\textbf{Sex DI}&\textbf{AUC} \\
\cmidrule(l){4-6} \cmidrule(l){7-8} \cmidrule(l){9-9}

\multirow{10}{*}{{COMPAS}}&Baseline& - & - & - & - & 0.837 & 1.086 & 0.654 \\
\cmidrule(l){2-3} \cmidrule(l){4-6} \cmidrule(l){7-8} \cmidrule(l){9-9}
&\multirow{4}{*}{{Union}}&Global DI& 1 & 0.0015 & 52.07 & 0.881 & 1.016 & 0.645 \\
&&Race DI& 2 & 0.0035 & 52.07 & \textbf{0.965} & 0.680 & 0.643 \\
&&Sex DI& 1 & 0.0015 & 2.41 & 0.853 & \textbf{1.000} & 0.649 \\
&&AUC& 2 & 0 & 34.95 & 0.874 & 0.805 & 0.662 \\
\cmidrule(l){2-3} \cmidrule(l){4-6} \cmidrule(l){7-8} \cmidrule(l){9-9}
&\multirow{4}{*}{{Intersec}}&Global DI& 2 & 0.0013 & 52.07 & \textit{\textbf{0.868}} & \textit{\textbf{0.999}} & \textit{\textbf{0.647}} \\
&&Race DI& 2 & 0.0035 & 52.07 & 0.885 & 0.966 & 0.643 \\
&&Sex DI& 2 & 0.0013 & 2.41 & 0.866 & 0.999 & 0.647 \\
&&AUC& 1 & 0 & 2.41 & 0.831 & 1.089 & \textbf{0.656} \\

\hline

&&&\textbf{N\_i}&\textbf{R}&\textbf{D}&\textbf{Race DI}&\textbf{Sex DI}&\textbf{AUC} \\
\cmidrule(l){4-6} \cmidrule(l){7-8} \cmidrule(l){9-9}

\multirow{10}{*}{{Adult}}&Baseline& - & - & - & - & 0.675 & 0.379 & \textbf{0.746} \\
\cmidrule(l){2-3} \cmidrule(l){4-6} \cmidrule(l){7-8} \cmidrule(l){9-9}
&\multirow{4}{*}{{Union}}&Global DI& 1 & 0.0154 & 0.88 & 0.713 & 0.397 & 0.739 \\
&&Race DI& 1 & 0.0154 & 8.63 & 0.723 & 0.363 & 0.730 \\
&&Sex DI& 1 & 0.0154 & 0.88 & 0.713 & 0.397 & 0.739 \\
&&AUC& 1 & 0 & 0.88 & 0.675 & 0.379 & \textbf{0.746} \\
\cmidrule(l){2-3} \cmidrule(l){4-6} \cmidrule(l){7-8} \cmidrule(l){9-9}
&\multirow{4}{*}{{Intersec}}&Global DI& 4 & 0.0154 & 6.73 & \textit{\textbf{1.006}} & \textit{\textbf{0.622}} & \textit{\textbf{0.709}} \\
&&Race DI& 4 & 0 & 8.63 & \textbf{0.995} & 0.619 & 0.709 \\
&&Sex DI& 4 & 0 & 6.73 & 0.983 & \textbf{0.622} & 0.710 \\
&&AUC& 1 & 0 & 0.88 & 0.675 & 0.379 & \textbf{0.746} \\

\hline
&&&\textbf{N\_i}&\textbf{R}&\textbf{D}&\textbf{Gender DI}&\textbf{Age DI}&\textbf{AUC} \\
\cmidrule(l){4-6} \cmidrule(l){7-8} \cmidrule(l){9-9}
\multirow{10}{*}{{German}}&Baseline& - & - & - & - & 1.028 & 1.295 & 0.633 \\
\cmidrule(l){2-3} \cmidrule(l){4-6} \cmidrule(l){7-8} \cmidrule(l){9-9}
&\multirow{4}{*}{{Union}}&Global DI& 1 & 0.7109 & 0.49 & 1.062 & \textbf{1.043} & 0.646 \\
&&Gender DI& 1 & 0.4383 & 0.22 & 1.015 & 1.269 & 0.643 \\
&&Age DI& 1 & 0.7109 & 0.49 & 1.062 & \textbf{1.043} & 0.646 \\
&&AUC& 1 & 0.7109 & 0.22 & 1.139 & 1.270 & 0.599 \\
\cmidrule(l){2-3} \cmidrule(l){4-6} \cmidrule(l){7-8} \cmidrule(l){9-9}
&\multirow{4}{*}{{Intersec}}&Global DI& 1 & 0.5810 & 0.44 & \textit{\textbf{0.984}} & \textit{\textbf{1.080}} & \textit{\textbf{0.618}} \\
&&Gender DI& 1 & 0.8257 & 0.36 & \textbf{1.000} & 1.419 & 0.633 \\
&&Age DI& 1 & 0.581 & 0.44 & 0.984 & 1.080 & 0.618 \\
&&AUC& 2 & 0.0363 & 0.44 & 1.229 & 1.229 & \textbf{0.649} \\
\hline

&&&\textbf{N\_i}&\textbf{R}&\textbf{D}&\textbf{Race DI}& - &\textbf{AUC} \\
\cmidrule(l){4-6} \cmidrule(l){7-7} \cmidrule(l){9-9}
\multirow{4}{*}{{Ricci}}&Baseline& - & - & - & - & 0.501 & - & 0.908 \\
\cmidrule(l){2-3} \cmidrule(l){4-6} \cmidrule(l){7-7} \cmidrule(l){9-9}
&\multirow{2}{*}{{Union}}&Race DI& 1 & 0 & 5.49 & \textit{\textbf{0.513}} & - & \textit{\textbf{0.916}} \\
&&AUC& 1 & 0 & 5.49 & \textit{\textbf{0.513}} & - & \textit{\textbf{0.916}} \\
\hline
\end{tabular}}
\label{compar_german_base2}
\end{center}
\end{table*}

All in all, according to the results shown in Figures \ref{compas-union} to \ref{adult-intersec}, as well as the overall results in Table \ref{compar_german_base2}, the following conclusions can be reached. 
\begin{itemize}
    \item The choice between union and intersection undersampling can greatly affect the minimisation of Disparate Impact, as shown by the results on Adult (where intersection was clearly the better choice). Given the fact that in the other cases both union and intersection fared similarly to one another, it would seem using intersection undersampling is the better choice overall.
    \item Choosing the right number of covered instances threshold is always very important, as this threshold defines the different result clusters in the Disparate Impact plots. One of the extremes amongst the percentiles chosen (being either the lowest or highest value) gave the best results in all datasets, although which one is the right choice depends on the dataset.
    \item The radius threshold shows the behaviour/shape of clusters defined by the number of covered instances threshold, whether it is mostly constant, increasing, decreasing, sigmoidal, etc. Depending on the location of the plotted cluster compared to the optimum Disparate Impact, different radius threshold values will be optimal.
    \item The density threshold gives different heights in the results cluster, so using different threshold values helps fine-tune Disparate Impact depending on the clusters location compared to the optimum Disparate Impact.
\end{itemize}

\subsection{Sampling methods comparison}\label{subsubsec:Fair-ONB-ana-compar}

This Section presents the results from the comparative study between Fair-ONB and FAWOS, two bias reduction sampling methods that are based on data neighbourhoods.

While the initial intention was to perform the comparison using all 4 datasets indicated in Section \ref{subsec:Fair-ONB-exp-conj}, the particularities of FAWOS would make using Adult unfeasible. The computational complexity of FAWOS, which is not prepared to work on datasets that big (even having already reduced its number of samples), is too high, as it uses 5NN for each sample multiple times per experiment. While the sample size could have been reduced again to include the dataset in this study, the drastic reduction necessary would likely distort the dataset characteristics.

In this study, accuracy was used to measure classifier performance and, since FAWOS uses the Adapted Disparate Impact, which is a simple transformation of Disparate Impact (see Equation \ref{ida}), that was the chosen Fairness metric.

The best results obtained with each method (and their optimal parameters) on the COMPAS, German and Ricci datasets are presented in Tables \ref{compar_metodos_compas} to \ref{compar_metodos_ricci}, respectively. As a reminder, FAWOS parameters are the weights for safe (S), boderline (B) and rare (R) samples (as outliers receive weight 0), as well as the oversampling factor (OF); meanwhile, the parameters of the Fair-ONB method are the number of covered instances (N\_i), radius (R) and density (D) thresholds.

\begin{table*}[!h]
\caption{Comparison, according to Adapted Disparate Impact and Accuracy between the best FAWOS 
and Fair-ONB results on the COMPAS dataset.}
\begin{center}
\scalebox{0.9}{
\begin{tabular}{cccccc}
\hline
\textbf{Method}&\textbf{Best Parameters}&\textbf{Race ADI}&\textbf{Sex ADI}&\textbf{Tot. Dist. Opti.}&\textbf{Accuracy} \\
\hline
FAWOS & Weights: \texttt{S=0, B=0.6, R=0.4, OF=0.8} & 0.474 & 0.758 & 0.768 & \textbf{0.780} \\
Fair-ONB Intersec. & Thresholds: \texttt{N\_i=3, R=0.00, D=2.47} & \textbf{0.886} & 0.972 & 0.142 & 0.771 \\
Fair-ONB Union & Thresholds: \texttt{N\_i=1, R=0.00, D=34.37} & 0.873 & \textbf{0.999} & \textbf{0.128} & 0.747 \\
\hline
\end{tabular}}
\label{compar_metodos_compas}
\end{center}
\end{table*}

\begin{table*}[!h]
\caption{Comparison, according to Adapted Disparate Impact and Accuracy between the best FAWOS 
and Fair-ONB results on the German dataset.}
\begin{center}
\scalebox{0.9}{
\begin{tabular}{cccccc}
\hline
\textbf{Method}&\textbf{Best Parameters}&\textbf{Age ADI}&\textbf{Gender ADI}&\textbf{Tot. Dist. Opti.}&\textbf{Accuracy} \\
\hline
FAWOS & Weights: \texttt{S=0, B=0.6, R=0.4, OF=1} & \textbf{0.820} & 0.910 & 0.270 & \textbf{0.686} \\
Fair-ONB Intersec. & Thresholds: \texttt{N\_i=1, R=0.57, D=0.36} & 0.796 & \textbf{0.942} & \textbf{0.262} & 0.672 \\
Fair-ONB Union & Thresholds: \texttt{N\_i=1, R=0.80, D=0.22} & 0.789 & 0.925 & 0.286 & 0.682 \\
\hline
\end{tabular}}
\label{compar_metodos_german}
\end{center}
\end{table*}

\begin{table*}[!h]
\caption{Comparison, according to Adapted Disparate Impact and Accuracy between the best FAWOS 
and Fair-ONB results on the Ricci dataset.}
\begin{center}
\scalebox{0.9}{
\begin{tabular}{cccc}
\hline
\textbf{Method}&\textbf{Best Parameters}&\textbf{Race ADI}&\textbf{Accuracy} \\
\hline
FAWOS & Weights: \texttt{S=0, B=0.6, R=0.4, OF=1.2} & 0.282 & 0.910   \\
Fair-ONB & Thresholds: \texttt{N\_i=1, R=0.04, D=5.52} & \textbf{0.467} & \textbf{0.917}   \\
\hline
\end{tabular}}
\label{compar_metodos_ricci}
\end{center}
\end{table*}

As can be observed, the Fair-ONB method obtains the best results according to Global Adapted Disparate Impact (as indicated by the lower total distance to the optimum Adapted Disparate Impact amongst protected features) and, very often, also the best Adapted Disparate Impact on the individual protected features. Given the usual inverse relationship between predictive performance and bias mitigation, it makes sense that FAWOS obtains better accuracy results.

\begin{figure}[!ht]
\centering

    \subfloat[]{\label{instancias-base}
    \includegraphics[width=.45\linewidth]{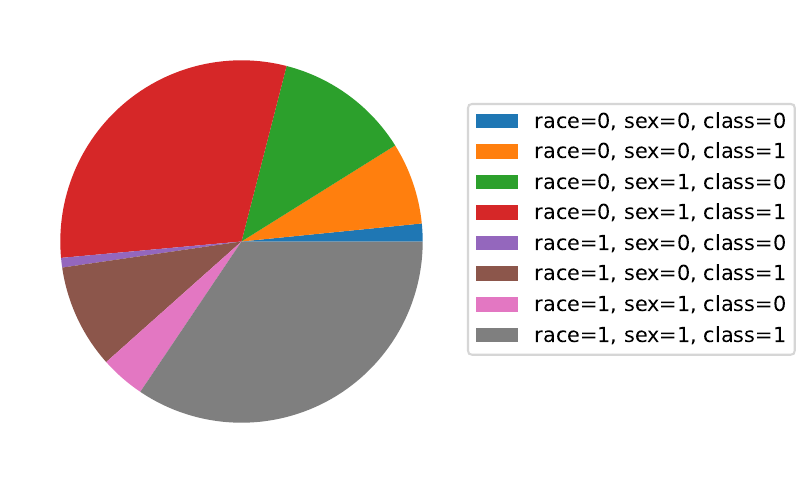}}
    \subfloat[]{\label{instancias-union}
    \includegraphics[width=.45\linewidth]{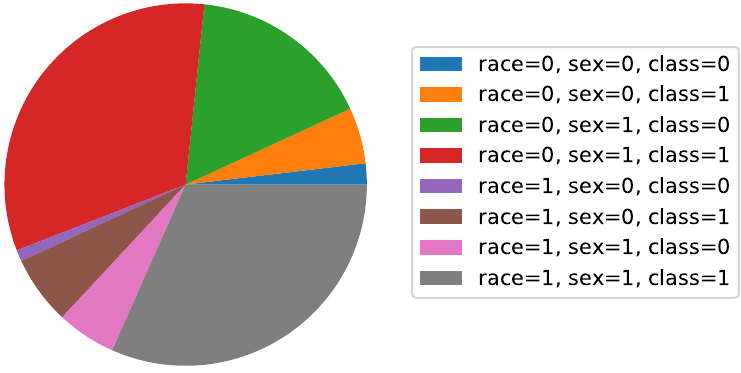}}

\caption{\label{fig:subgrupos compas}Example of COMPAS group proportions in the base case (\ref{instancias-base}) and with the best-performing union undersampling using the Fair-ONB method (\ref{instancias-union}), showing that achieving group balance was not the optimal strategy.}
\end{figure}

As a possible reason for this, Figure \ref{fig:subgrupos compas} shows two pie charts with the group distributions in the base COMPAS dataset and when the aforementioned best performing union undersampling is used. As can be observed, the best result is obtained without group balance, which is what most other methods (including FAWOS) would pursue, potentially giving Fair-ONB the edge. This also exemplifies the importance of correctly harnessing the morphology of protected groups towards modifying their proportions, as sampling in the most relevant parts of the dataset outweighed random sampling in group balancing (even if in FAWOS it was informed by neighbours according to 5NN).

\section{Concluding remarks} \label{concl}

This paper presents 
Fair-ONB, a new sampling method that employs guided undersampling in areas that are close to the decision boundaries, where models usually have problems classifying the different protected groups. The Fair-ONB method harnesses the morphology of groups to select samples for elimination, and unlike other random sampling techniques, its logic is more justifiable due to its empirical approach. 

As shown in the experimental study, the Fair-ONB method improves model Fairness while maintaining or improving classification performance, reflecting the usefulness of using subgroup ball coverage for undersampling instance selection. Moreover, the Fair-ONB method also achieves better bias reduction results than FAWOS, its neighbourhood-based competitor from the state of the art. 

Therefore, this paper has shown the usefulness of morphology and neighbourhood-based sampling approaches towards classifier Fairness, a niche that has not been thoroughly explored. In particular, the use of threshold variables and group coverage strategies, which give information on multiple morphological properties of the groups in a dataset, have been useful for the proper characterisation of problematic overlap areas. Thus, the hypothesis that using data morphology for undersampling instance selection can be a very useful strategy is confirmed.

Regarding future work lines, combining morphology-based strategies with other preprocessing techniques could lead to attaining higher fairness levels. Using a bigger scope and finding synergies could allow for the implementation of more in-depth preprocessing techniques.

\section*{Acknowledgments}

This work has received funding from the Spanish Ministry of Science and Technology under project PID2020-119478GB-I00, including European Regional Development Funds. It is also partially supported by the I+D+i project granted by C-ING-250-UGR23 co-funded by ``Consejería de Universidad, Investigación e Innovación'' and the European Union related to FEDER Andalucía Program 2021-27. 

\bibliography{references}
\bibliographystyle{spmpsci}

\vspace{12pt}

\appendix

\end{document}